\documentclass[11pt]{article}

\usepackage[final]{acl}

\usepackage{times}
\usepackage{latexsym}

\usepackage[T1]{fontenc}

\usepackage[utf8]{inputenc}

\usepackage{microtype}

\usepackage{inconsolata}

\usepackage{graphicx}

\usepackage{amssymb}
\usepackage{amsmath}
\usepackage{subcaption}
\usepackage{booktabs} 
\usepackage{multirow}
\usepackage{algorithm}          
\usepackage{algorithmic}        
\usepackage[table]{xcolor}
\usepackage[colorinlistoftodos]{todonotes}
\title{ANI: Adaptive Numerical Injection for \\
  Unifying Semantic and Arithmetic Representations in Numerical Reasoning}

\author{Jinsung Jeon \\
  KAIST InnoCORE LLM \\
  Seoul National University \\
  Korea University \\
  \texttt{jinsungjeon@korea.ac.kr} \\\And
    Seung-won Hwang\thanks{Corresponding Author} \\
  Seoul National University \\
  \texttt{seungwonh@snu.ac.kr} \\}

\begin{document}
\maketitle
\begin{abstract}
    Precise numerical reasoning with Large Language Models (LLMs) is essential for expanding their applicability to complex real-world tasks. 
    However, text-based tokenization often fragments numbers, significantly hindering precise arithmetic reasoning.
    Meanwhile, numerical embeddings, despite arithmetic precision, rely on context-agnostic substitution that disregards the semantic role of numbers as identifiers.
    To combine the complementary strengths, we propose \textbf{ANI (Adaptive Numerical Injection)}, a hybrid framework that governs the selective injection of numerical features based on the semantic context.
    By employing a context-aware gating mechanism, we selectively inject numerical embeddings (specifically FoNE) into the latent space, explicitly preserving nominal identifiers while enhancing quantitative operands.
    Through extensive evaluations across various LLMs, we demonstrate that ANI enhances MATH performance by 9.5 points over the official reference model, while maintaining robust performance on general linguistic benchmarks\footnote{https://github.com/Jinsung-Jeon/ANI\_EMNLP}.
\end{abstract}

\section{Introduction}

\begin{figure}[t]
  \vskip 0.2in
  \begin{center}
    \centerline{\includegraphics[width=\columnwidth]{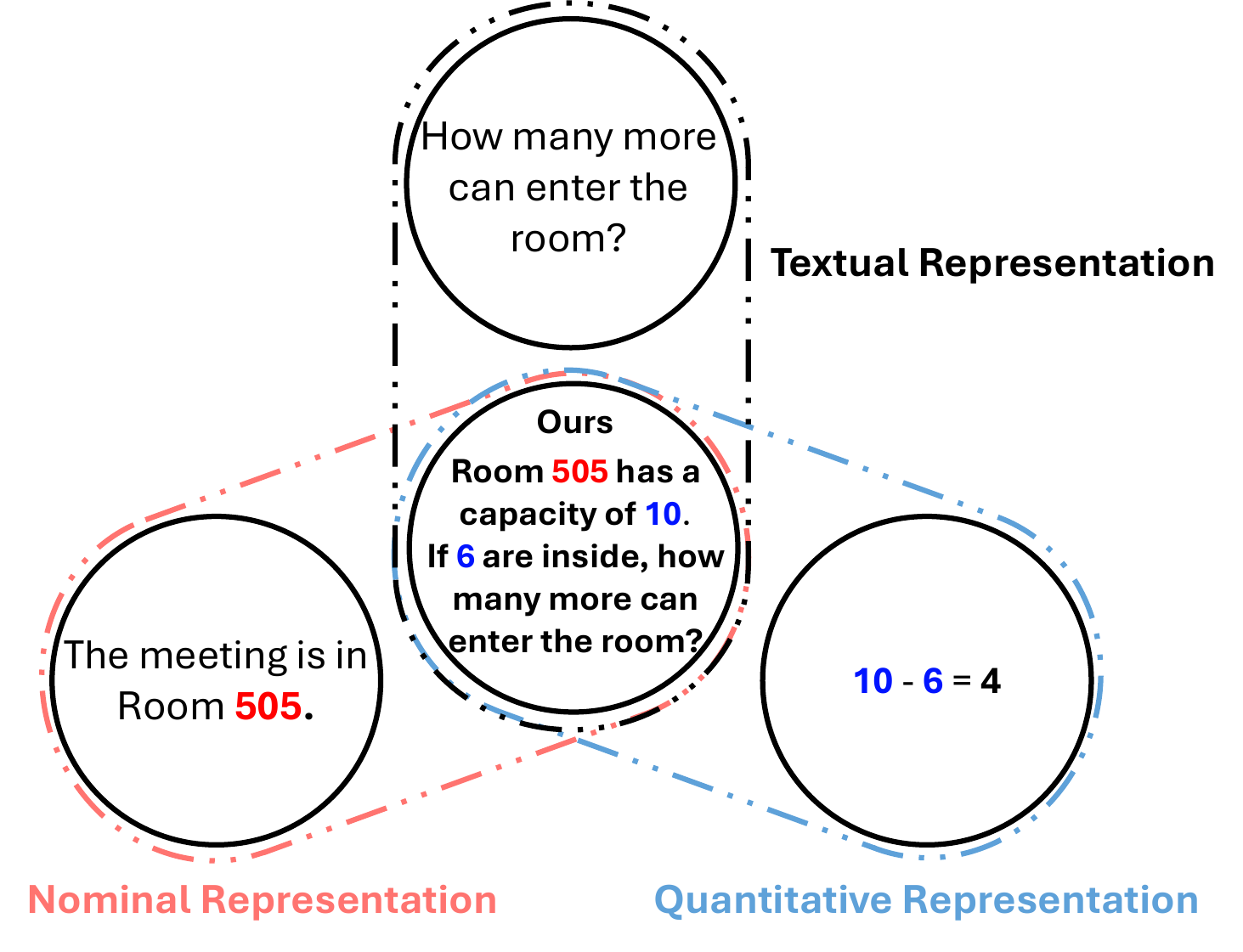}}
    \vspace{-1em}
    \caption{
      \textbf{Overview of the proposed adaptive representation.} Unlike existing methods that rely exclusively on uniform textual representations, our framework dynamically integrates a unified architecture by explicitly distinguishing between nominal representations and quantitative representations. 
    }
    \label{fig:overall}
  \end{center}
  \vspace{-3.5em}
\end{figure}

Large Language Models (LLMs) have demonstrated remarkable capabilities across a wide range of natural language processing tasks~\cite{brown2020language,achiam2023gpt,grattafiori2024llama,yang2025qwen3}. Despite this success, their performance in precise numerical reasoning remains a significant challenge due to the inherent limitations of discrete tokenization. Traditional architectures typically treat numbers as fragmented sequences of text tokens (e.g., sub-words), relying on linguistic probability distributions to predict the next numerical digit~\cite{wallace2019nlp, nogueira2021investigating}. Consequently, the model struggles to capture the numerical magnitude and mathematical properties, often generating results based on surface-level text patterns rather than rigorous logical computation~\cite{dziri2023faith}.

To enhance arithmetic precision, recent works have introduced continuous numerical 
embeddings~\cite{golkar2023xval,schwartz2024numerologic,zhou2025fone}. However, these methods often fall into the trap of \textit{Context-Agnostic Substitution}, or indiscriminately replacing with a numerical embedding. While mathematically precise for pure arithmetic (e.g., addition, subtraction), this naive substitution leads to semantic erasure for tokens, working as nominal identifiers.

We argue that this deficiency presents a critical bottleneck in numerical reasoning, ranging from financial analysis to technical documentation.
Figure~\ref{fig:overall} illustrates real-life numerical reasoning that require a unified representation of numerical data embedded within natural language contexts. For example, in the sentence "Room 505 has a capacity of 10", the number "\textcolor{red}{505}" serves as a \textbf{nominal identifier} (a room ID) that should be processed as a textual entity. In contrast, "\textcolor{blue}{10}" is a \textbf{quantitative value} intended for arithmetic reasoning.
Standard tokenizers would treat all digits uniformly, potentially misinterpreting a non-calculable identifier like "Room ID 505" as a summable quantity, or conversely, failing to capture numerical magnitudes by treating them as mere text tokens. Consequently, reasoning fails when models cannot distinguish between numeric labels and quantitative values.


To resolve this dilemma, we propose \textbf{ANI (Adaptive Numerical Injection)}.
Unlike static approaches, ANI does not blindly overwrite representations. Instead, it probes the backbone model's latent states to interpret the role of a number, either as nominal IDs or quantitative operands. By deriving a differentiable binary decision via this internal probing, ANI \textit{adaptively injects} mathematical structures only when the context explicitly demands arithmetic reasoning. This balanced approach allows the model to preserve its linguistic integrity while selectively enhancing numerical precision. Consequently, ANI excels not only in numerical reasoning where numbers and text coexist but also remains robust in text- or numeric-only scenarios.
\noindent \textbf{Contributions.} Our contributions are as follows: 
\begin{itemize}
    \item We propose ANI, which dynamically adapts between nominal identifiers and quantitative representations by injecting mathematical topology only when necessary.
    \item ANI enables retaining pre-trained reasoning and selectively boosting arithmetic precision, which we identified as a key bottleneck in numerical reasoning.
    \item We validate the robustness of ANI through extensive evaluations across diverse model families and scales. 
    Notably, ANI achieves a 9.5-point improvement on the MATH benchmark while incurring a marginal performance drop of only 1.1\% on general linguistic tasks in Qwen3-8B.
\end{itemize}

\section{Related work and Preliminaries}\label{sec:related_work}

\paragraph{Arithmetic Reasoning in LLMs.}
While LLMs have achieved remarkable success in natural language understanding, they persistently struggle with arithmetic reasoning and precise calculation. Early hypotheses relied on scaling laws~\cite{kaplan2020scaling, wei2022emergent}, assuming that larger models would naturally acquire mathematical proficiency. However, empirical studies show that standard Transformers~\cite{vaswani2017attention} often fail to perform consistent arithmetic operations due to the fragmented nature of subword tokenization mechanisms, such as Byte-Pair Encoding~\cite{sennrich2016neural}. Since these algorithms prioritize text compression, they often split numbers into inconsistent chunks (e.g., "1234" $\rightarrow$ "12", "34"), disrupting semantics necessary for calculation.

To address these architectural deficiencies, current state-of-the-art methods predominantly employ inference-time strategies such as Chain-of-Thought (CoT) prompting~\cite{wei2022chain,tang2026multiplex}, which decomposes complex problems into intermediate natural language steps. More recently, program-aided approaches have gained prominence, where models generate and execute programming code (e.g., Python) to offload precise calculations to external interpreters~\cite{chen2022program,gao2023pal}. While effective, these methods mitigate the limitations externally rather than the root cause: the suboptimal internal representation of numbers within the model itself.

\begin{figure*}[t]
  \begin{center}
    \centerline{\includegraphics[width=0.9\textwidth]{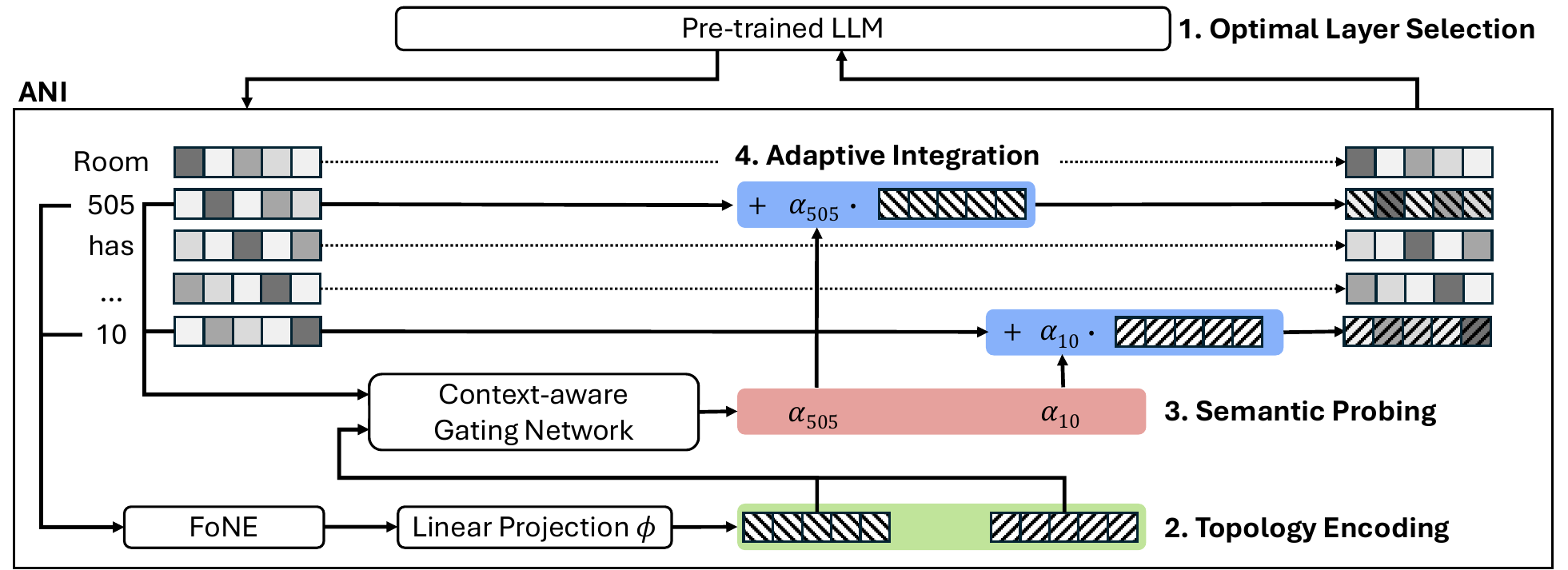}}
    \caption{\textbf{The overall architecture of ANI.} Numerical embeddings are generated from raw scalar values and mapped to the model's hidden dimension via a Linear Projection. Simultaneously, the Context-aware Gating Network computes a scalar coefficient ($\alpha$) based on both the hidden state of each numerical token and its numerical embedding. The final representation is obtained by adding the projected numerical embedding, scaled by $\alpha$, to the original hidden state. Note that non-numerical tokens bypass this module to preserve their linguistic features.}
    \label{fig:ANI}
  \end{center}
  \vspace{-2.5em}
\end{figure*}

\paragraph{Number Representations.}
Recent literature highlights that LLMs inherently possess highly accurate internal representations of numbers that remain stable throughout internal processing~\cite{kadlvcik2025pre}.
Despite these accurate internal representations, structural fragmentation caused by subword tokenization severely hinders complex arithmetic. 
As empirically demonstrated in Section~\ref{Sec:Additional}, this fragmentation leads to calculation failures on longer sequences.

To address this, early attempts to improve numeracy in LLMs often relied on digit-wise representations or naive additive fusion ($h_{num} \leftarrow h_{num} +e_{num}$). In these approaches, numerical features are simply summed with textual embeddings at the input layer. However, as analyzed in recent literature~\cite{zhou2025fone}, such indiscriminate addition leads to signal interference, where the quantitative and semantic signals become entangled and indistinguishable within the same vector space. This interference often leads to catastrophic forgetting, as the model's pre-trained linguistic manifold is disrupted by raw numerical noise.

To mitigate such entanglement, state-of-the-art methods such as xVal~\cite{golkar2023xval} and FoNE~\cite{zhou2025fone} adopt a static substitution strategy. By replacing textual tokens with precise single-token embeddings ($h_{num} \leftarrow e_{num}$), these methods successfully capture mathematical properties like magnitude and modular arithmetic without the noise of naive addition. Yet, this creates a new structural dilemma: \textit{semantic erasure}. Since the substitution is static and context-agnostic, numbers serving as nominal identifiers (e.g., "Room 505") lose their linguistic identity, leading to performance degradation in tasks where numbers and text must coexist. ANI reconciles this conflict by transitioning from context-agnostic substitution back to an additive scheme, but one that is strictly governed by an `adaptive' and `context-aware' gating process ($h_{num} \leftarrow h_{num} + \alpha \cdot e_{num}$).

\section{Proposed Method}

In this section, we introduce \textbf{ANI (Adaptive Numerical Injection)}, a hybrid architecture designed to adaptively inject numerical embeddings into the hidden states of pre-trained LLMs.

\subsection{Problem Formulation}
\label{sec:problem_formulation}
Existing numerical integration strategies suffer from a trade-off between semantic preservation and arithmetic precision. While \textit{additive fusion} preserves semantics, it induces signal entanglement with the linguistic manifold, hindering precise calculation. Conversely, \textit{static substitution} ensures precision but triggers semantic erasure, stripping numbers of their nominal identity. Both failure modes stem from context-blind intervention at the input level, which forces a functional interpretation before the model gains sufficient contextual depth to distinguish between quantitative operands and nominal identifiers.

To resolve this structural dilemma, we propose ANI, a framework designed to function as a context-aware semantic probe. Instead of premature intervention at the input level, ANI shifts the numerical injection to the optimal layer where the model has developed sufficient contextual maturity. By leveraging these contextually enriched representations, ANI enables the selective integration of numerical features, ensuring that mathematical structures are reinforced only when necessitated by the semantic context. This approach allows the model to enhance its numerical precision while fundamentally preserving its pre-trained linguistic integrity.

\subsection{Overall Workflow}
\label{sec:workflow}
Based on the formulation above, we implement ANI as a plug-and-play module that intervenes in the intermediate layers of the backbone LLM. As illustrated in Figure~\ref{fig:ANI}, the framework operates through the following four phases:
\begin{enumerate}
    \item \textbf{Optimal Layer Selection:} We select an intermediate layer $l$ where the hidden states have developed sufficient contextual maturity to resolve numerical ambiguity.
    \item \textbf{Topology Encoding (Numerical Path):} Raw scalars are transformed into numerical embeddings and aligned with the latent space through a Linear Projection.
    \item \textbf{Semantic Probing (Gating Path):} A Gating Network computes a discrete gating coefficient $\alpha$ by interpreting the terminal token's hidden state to distinguish between nominal and quantitative roles.
    \item \textbf{Adaptive Integration:} The aligned embedding, scaled by $\alpha$, is adaptively injected via gated residual connection, while non-numerical tokens bypass the module to ensure zero interference.
\end{enumerate}
Based upon this workflow, the following subsections detail the implementation of our framework.

\begin{figure}[t]
    \centering
    \includegraphics[width=1.0\linewidth]{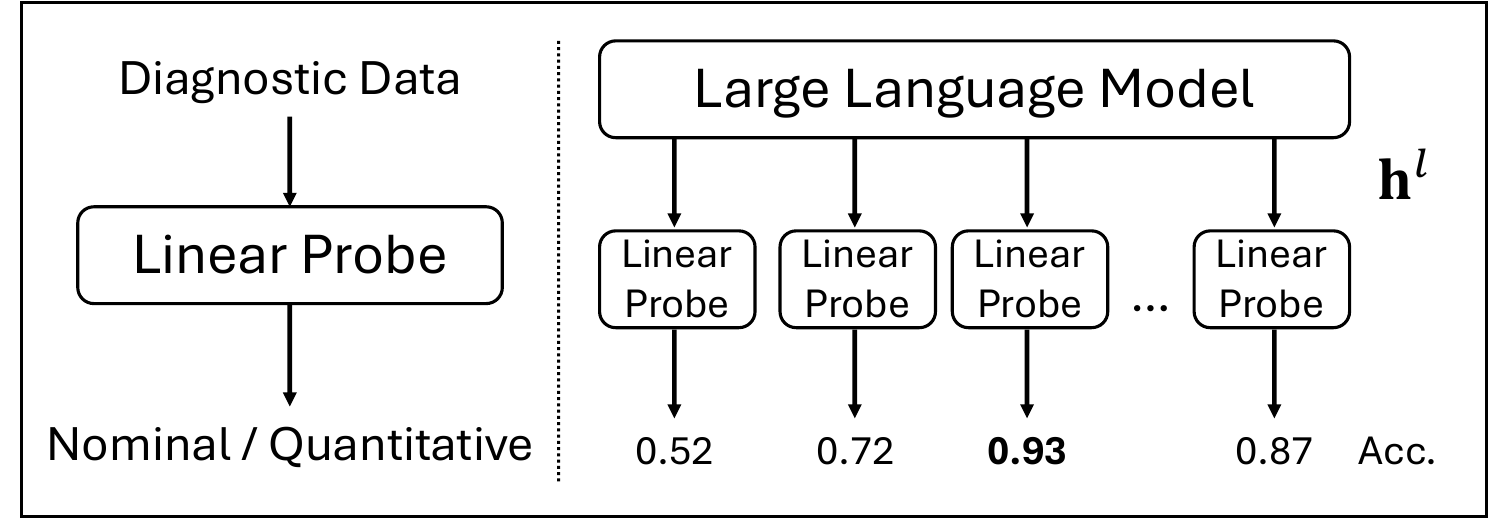}
    \vspace{-3mm} 
    \caption{\textbf{Linear probing analysis for optimal layer selection.} We evaluate the degree of numerical disambiguation across layers using a diagnostic dataset. The peak in classification accuracy identifies the layer with maximum contextual maturity, which serves as the optimal point for adaptive numerical injection.}
    \label{fig:optimal_layer_selection}
    \vspace{-1.5em} 
\end{figure}

\subsection{Optimal Layer Selection and Topology Encoding}
\label{sec:layer_and_encoding}
ANI is designed to integrate numerical information at a specific intermediate depth where the hidden representations have attained sufficient contextual maturity to resolve the functional ambiguity between nominal and quantitative roles. 

\paragraph{Optimal Layer Selection.} 
To address the limitations of context-agnostic input-level substitution, we identify an optimal injection layer $l$ that exhibits sufficient \textit{contextual maturity} via linear probing analysis (cf. Figure~\ref{fig:optimal_layer_selection}). Specifically, we train layer-wise linear probes on a balanced diagnostic dataset to classify the functional roles of numerical tokens (nominal vs. quantitative). The classification accuracy serves as a proxy for the degree of numerical disambiguation achieved at each depth. We designate the layer where this accuracy peaks as the optimal locus for numerical injection, allowing the gating mechanism to leverage contextually enriched representations $\mathbf{h}^{l} \in \mathbb{R}^{d_{model}}$, where $d_{model}$ represents the hidden dimension of the backbone model. For our experiments with Qwen3-8B, we select $l=24$ as the injection point based on this protocol; a detailed empirical validation of this selection is provided in Section~\ref{Sec:Additional}.

\paragraph{Topology Encoding (Numerical Path)}
Once the injection layer is identified, we ensure that the numerical information is represented in a format that preserves its intrinsic mathematical properties. To address the discreteness of subword tokenization, we adopt Fourier Number Embedding~\cite{zhou2025fone} as our base encoder to represent each numerical value as a single-token entity. Given a scalar value $x$, Fourier Number Embedding (FoNE) maps it to a high-dimensional feature vector $\phi(x) \in \mathbb{R}^{d_{fourier}}$ using periodic activation functions, effectively capturing both magnitude and fractional details.

Crucially, the raw Fourier features $\phi(x)$ exist in a mathematical manifold distinct from the pre-trained LLM's semantic space. Direct usage of these features creates a \textit{distributional mismatch}. To bridge this gap, we introduce a learnable Linear Projection $W_P \in \mathbb{R}^{d_{model} \times d_{fourier}}$. This projection serves a dual purpose: (i) dimensionality matching, and (ii) manifold alignment, transforming the raw mathematical features into a representation $e_{num}$ compatible with the backbone's latent space:
\begin{equation}
    e_{num} = W_P \cdot \phi(x).
\end{equation}
Unlike naive approaches that employ digit-wise injection, which often fail to capture the compositional structure of numbers, our approach generates a single unified numerical embedding for the entire number entity. This ensures that the global numerical value is preserved and injected strictly at the terminal token, enabling a compact and contextually grounded integration as detailed in the subsequent gating phase.

\begin{figure}[t]
  \begin{center} 
    \centerline{\includegraphics[width=\columnwidth]{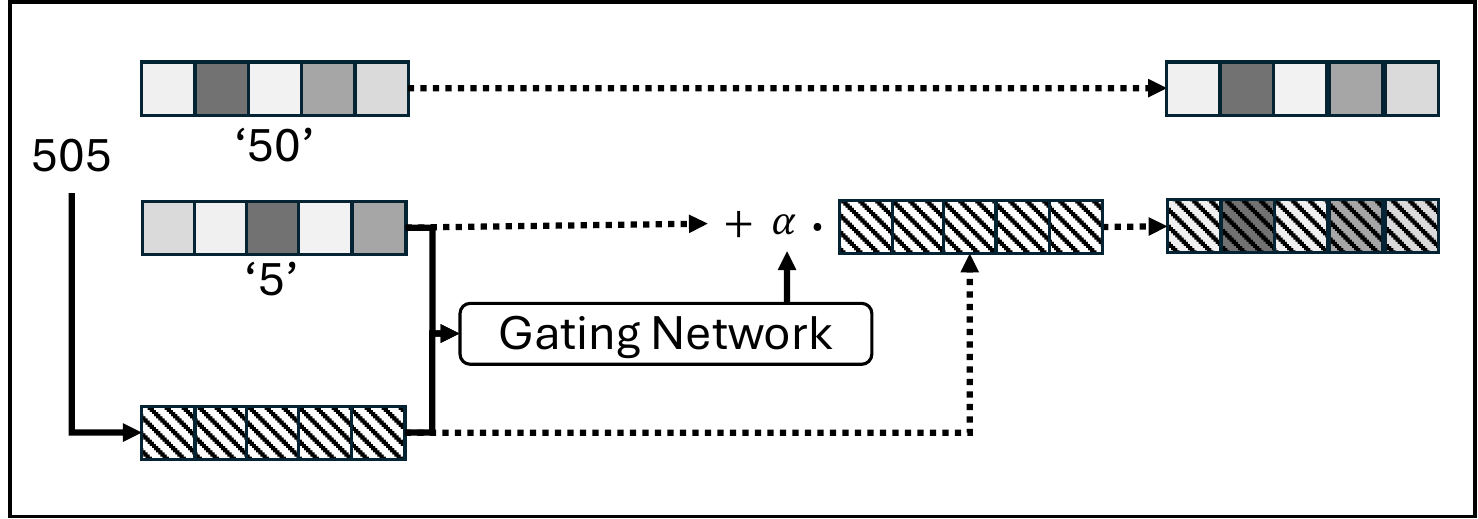}}
    \vspace{2mm} 
    \caption{
      \textbf{Gating mechanism and last token injection.} ANI selectively integrates numerical embeddings into the terminal sub-token's hidden state via an adaptive gate $\alpha$. This mechanism ensures that mathematical structures are reinforced only when semantically required, preserving the model's linguistic integrity.
    }
    \label{fig:gating_mechanism}
  \end{center}
  \vspace{-3em} 
\end{figure}

\subsection{Semantic Probing and Adaptive Integration} \label{sec:gating}
As shown in Figure~\ref{fig:gating_mechanism}, ANI uses a context-aware gating mechanism to determine the selective activation of numerical features at the last sub-token.

\paragraph{Semantic Probing (Gating Path)} To achieve a seamless integration that preserves linguistic context while enhancing arithmetic precision, we address two design challenges: (i) \textbf{Target Locus (\textit{Where})}, the optimal aggregation point to minimize signal redundancy, and (ii) \textbf{Functional Selection (\textit{How})}, the mechanism to discern the necessity of mathematical intervention.

First, we determine \textbf{\textit{where}} to inject the numerical features by resolving the granularity mismatch inherent in subword tokenization. While a numerical entity possesses a unified mathematical identity, it is often fragmented into a sequence of sub-tokens $S=\{t_1, \dots, t_L\}$ in the textual space. To bridge this gap, we consider two integration strategies: 
(i) \textbf{Digit-wise Injection}: This approach broadcasts the modulated numerical embedding to every sub-token $t_i \in S$. We argue that even with adaptive modulation, this strategy remains suboptimal as it induces $L$-fold signal interference. Such repetitive intervention leads to feature entanglement, where numerical noise cumulatively disrupts the established representation space of the backbone. This dense redundancy effectively overwhelms the local semantic context with redundant mathematical topology, negating the benefits of adaptive injection.
(ii) \textbf{Terminal Token Injection} (Ours): Instead of broadcasting, we leverage the causal nature of Decoder-only Transformers~\cite{vaswani2017attention}. 
Although attention weights can theoretically distribute across all preceding tokens, the last token $t_L$ functionally serves as an information bottleneck that aggregates this context.
Consequently, we strictly inject $e_{num}$ into the hidden state $h_{t_{L}}^{(l)}$. 
This sparse intervention ensures that the holistic numerical topology modulates the representation precisely once, effectively safeguarding the model’s linguistic integrity while minimizing the risk of degradation associated with dense feature injections. The empirical superiority of this sparse injection approach over the digit-wise alternative is further substantiated by our ablation results, as detailed in Table~\ref{tab:ablation}.

Next, we address \textbf{\textit{how}} to derive a discrete gating decision through context-aware semantic probing. To ensure that numerical enhancement does not compromise pre-established fluency, our \textbf{context-aware gating network} operates as a continuous scaler during training, which is explicitly regularized to converge into a binary selector. By treating the contextual hidden state $\mathbf{h}^{(l)}_{t_{L}}$ as the \textit{subject} that evaluates the mathematical necessity of the numerical \textit{object} $\mathbf{e}_{num}$, ANI selectively filters out potential interference. Our gating MLP computes a raw logit $z_{num}$ mapped via a standard sigmoid function:
\begin{equation}
z_{num} = MLP_{gate}([\mathbf{h}^{(l)}_{t_{L}}; \mathbf{e}_{num}]),
\end{equation}
\vspace{-1em}
\begin{equation}
\alpha_{num} = \sigma(z_{num}),
\end{equation}
where $\alpha_{num} \in (0, 1)$ dictates the injection ratio. To induce discreteness and push the continuous scale toward a deterministic 'on/off' switch without breaking differentiability, we introduce a binarization penalty to the standard Next-Token Prediction (NTP) objective:
\begin{equation}
\mathcal{L} = \mathcal{L}_{\mathrm{CE}} + \lambda \mathbb{E}[\alpha_{num}(1-\alpha_{num})],
\end{equation}
where $\lambda$ (e.g., $0.1$) controls the regularization strength. This penalty is maximized at $\alpha_{num} = 0.5$ and minimized as $\alpha_{num} \to 0$ or $1$. During inference, to guarantee absolute zero-interference for nominal identifiers, we bypass the continuous scale and apply a hard threshold:
\begin{equation}
\alpha_{num} = \mathbb{I}(\sigma(z_{num}) > 0.5).
\end{equation}

\paragraph{Adaptive Integration.}
The final integration of numerical features is realized through an additive augmentation, where the gate $\alpha_{num}$ dictates the operational regime of the model:
\begin{equation}
    \tilde{\mathbf{h}}^{(l)}_{t_L} = \underbrace{\mathbf{h}^{(l)}_{t_L}}_{\text{Base Semantic}} + \underbrace{\alpha_{num} \cdot \mathbf{e}_{num}.}_{\text{Conditional Enhancement}}
\end{equation}
As defined by the gating decision, this formulation enables two distinct functional modes: (i) \textbf{Active Injection} ($\alpha_{num} = 1$), where the numerical path is activated for quantitative operands to reinforce arithmetic reasoning, and (ii) \textbf{Neutral Preservation} ($\alpha_{num} = 0$), where the gate remains closed for nominal identifiers to ensure zero-interference.

In contrast to traditional mixing or substitution approaches—where enhancing one feature necessitates the degradation of another in a zero-sum trade-off—our additive scheme maintains the base semantic vector $\mathbf{h}^{(l)}_{t_L}$ as an immutable foundation. This architectural choice provides the flexibility to inject precise mathematical topology as a calibrative refinement while filtering out potential numerical noise. Crucially, all non-numerical tokens bypass this module entirely, ensuring that the pre-trained linguistic manifold remains uncompromised for the vast majority of the sequence. This selective assimilation allows ANI to maximize mathematical precision only when functionally necessary, effectively mitigating the risk of semantic erasure and catastrophic forgetting.

\subsection{Training Strategy and Objective} \label{sec:training}
To integrate the ANI module while preserving the pre-trained linguistic manifold, we utilize Quantized Low-Rank Adaptation (QLoRA)~\cite{dettmers2023qlora} to update only the lightweight adapters and the ANI module. Rather than relying on stochastic sampling estimators, the model is optimized using the standard Next-Token Prediction (NTP) objective augmented with our binarization penalty $\lambda \mathbb{E}[\alpha_{num}(1-\alpha_{num})]$. This explicit regularization ensures stable convergence from a continuous scaling state toward a deterministic hard-selection regime. This unified protocol allows ANI to optimally utilize numerical topology while strictly safeguarding the model's general-purpose capabilities. The detailed formulation and the full execution procedure are deferred to Appendix Algorithm~\ref{alg:ani_forward}.


\section{Experiments}
In this section, we empirically evaluate ANI's capability to enhance mathematical reasoning while preserving general-purpose linguistic knowledge. Detailed setup is found in the Appendix~\ref{sec:detailed}.

\subsection{Experimental Setup}
\paragraph{Datasets and Metrics.} We evaluate ANI on two core dimensions: (1) Numerical Reasoning via MATH~\cite{hendrycks2021measuring} and GSM8K~\cite{cobbe2021gsm8k}, and (2) General Capabilities via MMLU~\cite{hendryckstest2021} to monitor potential semantic erasure. For all tasks, we report Accuracy (Acc) via a Chain-of-Thought protocol to ensure rigorous logical reasoning. For full reproducibility, our complete prompt templates are provided in Appendix~\ref{app:evalprompt}.

\paragraph{Baselines.} To evaluate the efficacy of our context-aware integration, we compare ANI against three categories of state-of-the-art methodologies: (1) \textbf{Text-based models}, including the official Reference and Standard SFT; (2) \textbf{Input-level Integration}, which modifies representations at the input stage via context-agnostic substitution—such as xVal \cite{golkar2023xval} and FoNE \cite{zhou2025fone}—or additive fusion (Digit-wise); and (3) \textbf{In-context methods} such as NumeroLogic \cite{schwartz2024numerologic}, which inject structural numerical metadata directly into the input sequence.

\begin{table}[t]
\begin{center}
\begin{small}
\begin{sc}
\resizebox{\columnwidth}{!}{
\begin{tabular}{lccc}
\toprule
\multirow{2}{*}{Model} & \multicolumn{2}{c}{Mathematical Reasoning} & General \\ 
\cmidrule(lr){2-3} \cmidrule(lr){4-4}
 & MATH (Acc) & GSM8K (Acc) & MMLU (Acc) \\ 
\midrule
\multicolumn{4}{l}{\textit{Reference Models (Official)}} \\
Qwen3-8B  & 49.28 & 89.84 &  73.75 \\
\midrule
\multicolumn{4}{l}{\textit{Standard Fine-tuning}} \\
Qwen3-8B & 55.18 & 87.79 & 72.24 \\ 
\midrule
\multicolumn{4}{l}{\textit{Controlled Experiments (Methods)}} \\
Digit-wise & 56.44 & 88.17 & 72.56 \\ 
xVal & 58.32 & \textbf{90.83} & 72.50 \\ 
FoNE & 55.08 & 89.61 & 72.30 \\ 
NumeroLogic & 47.56 & 42.99 & 72.45 \\ 
\midrule
\rowcolor{gray!15}
\textbf{ANI (Ours)} & \textbf{58.76} & 90.14 & \textbf{72.67} \\ 
\bottomrule
\end{tabular}
} 
\end{sc}
\end{small}
\end{center}
\vspace{-1em}
\caption{\textbf{Main Results on Qwen3-8B.} We compare our proposed ANI method with an official baseline and controlled baselines. \textbf{Bold} indicates the best performance among models trained on the same dataset.}
\label{tab:main_results}
\end{table}

\paragraph{Implementation Details.} We evaluate the Qwen3~\cite{yang2025qwen3} (4B, 8B, 32B) family as our primary backbones, with further validation on Llama-3.1~\cite{grattafiori2024llama} and Mistral. To train the gating mechanism’s ability to distinguish between nominal identifiers and quantitative operands, we construct a composite diagnostic dataset by pairing OpenMathInstruct-2~\cite{toshniwal2024openmathinstruct} (mathematical context) with Magpie~\cite{xu2024magpie} (linguistic identifiers). All models are optimized via QLoRA.

\subsection{Main Results} 
In this section, we empirically evaluate ANI by investigating its impact on specialized reasoning and general linguistic stability. Specifically, we address (RQ1) whether adaptive injection improves complex arithmetic performance and (RQ2) whether context-aware integration effectively mitigates the trade-off between specialized precision and general linguistic integrity.

\paragraph{RQ1: Impact on Mathematical Reasoning.}
To evaluate the arithmetic improvement enabled by ANI, we analyze the performance results summarized in Table~\ref{tab:main_results}. 
The results demonstrate that incorporating numerical embeddings significantly facilitates complex reasoning by providing the necessary mathematical topology. 

On the challenging MATH benchmark, ANI achieves an accuracy of 58.76\%, representing a substantial improvement of nearly 9.5 points over the official pre-trained reference model.
Notably, ANI surpasses not only the Standard SFT baseline but also existing context-agnostic strategies such as xVal and FoNE. 
Furthermore, ANI's advantage emerges on highly complex tasks. As detailed in our difficulty-wise analysis (Appendix~\ref{app:confidence_intervals}), it demonstrates robust gains on the hardest problems (MATH Levels 4 and 5) over input-level baselines.
This confirms that our adaptive integration is highly effective for high-level arithmetic where indiscriminate substitution falters.
While xVal shows a marginal lead on GSM8K, ANI remains highly competitive at 90.14\%, effectively outperforming the Reference model and significantly exceeding in-context methods like NumeroLogic. These findings indicate that ANI provides precise numerical features that reinforce the model's arithmetic reasoning without obstructing its fundamental reasoning path.

\begin{figure}[t]
    \centering
    \includegraphics[width=0.99\linewidth]{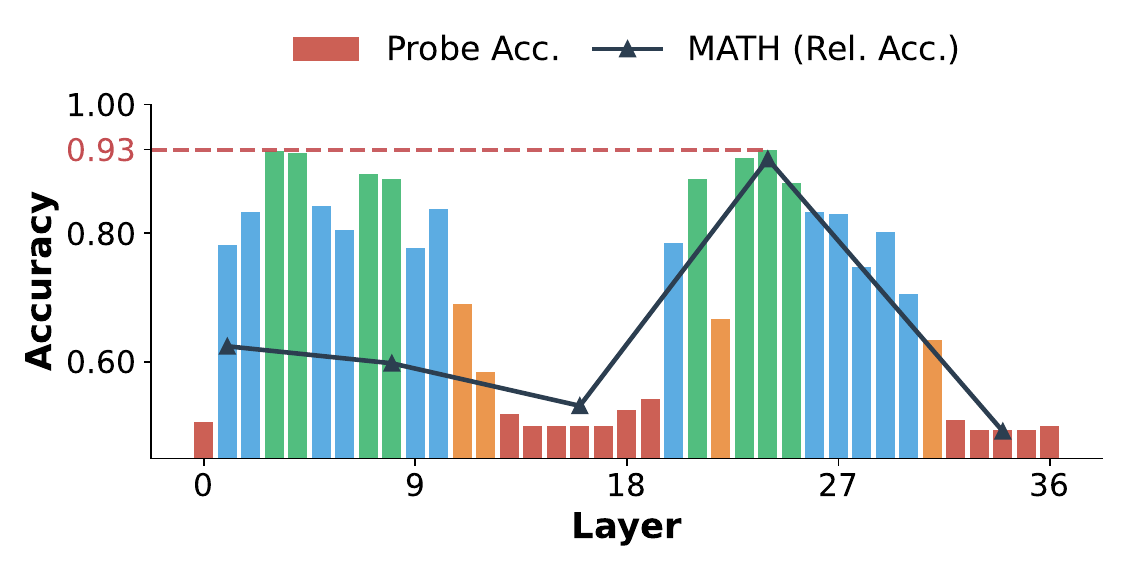}
    \caption{\textbf{Layer Sensitivity Analysis on MATH.} Performance is measured by relative accuracy (Rel. Acc.) on the MATH benchmark. The linear probe reaches a peak accuracy of 93\% at Layer 24.}
    \label{fig:layer_sensitivity}
\end{figure}

\paragraph{RQ2: Mitigation of General Language Understanding.} 
To investigate whether ANI successfully preserves pre-trained linguistic knowledge, we monitor performance on the MMLU benchmark, with the comparative results reported in Table 1. The results highlight a critical advantage of our gating architecture: while Standard SFT suffers a notable degradation from 73.75\% to 72.24\%, ANI successfully recovers a significant portion of this loss, maintaining a robust 72.67\% accuracy. 
In comparison, context-agnostic input-level baselines, such as FoNE and xVal, suffer from severe catastrophic forgetting.
This performance delta confirms that our Gating Network effectively distinguishes between computational and non-computational contexts.
By suppressing unnecessary numerical noise, ANI effectively mitigates the semantic drift typical of these baselines, thereby safeguarding the model's original representation manifold.

\subsection{Analysis of Adaptive Injection}
\label{Sec:Additional}
In this section, we conduct a series of diagnostic experiments to interpret the internal mechanism of ANI, validating our hypothesis regarding optimal feature integration and tool-routing synergy.

\paragraph{Robustness to Token Fragmentation.}
While LLMs possess a latent capacity to understand numerical semantics, numerical injection specifically addresses the critical functional bottleneck of information loss caused by token fragmentation in long numerical sequences. To empirically validate the necessity of explicit numerical injection, we conducted an N-Digit Arithmetic Scaling experiment using 100 evaluation problems for each digit length to measure exact match accuracy.

\begin{table}[t]
    \centering
    \resizebox{\linewidth}{!}{
    \begin{tabular}{lcccccc}
        \toprule
        \textbf{Method} & \textbf{2-digit} & \textbf{4-digit} & \textbf{6-digit} & \textbf{8-digit} & \textbf{10-digit} & \textbf{12-digit} \\
        \midrule
        Pre-trained (Qwen3-8B) & 75.0 & 48.0 & 6.0 & 0.0 & 0.0 & 0.0 \\
        Baseline (SFT) & \textbf{100.0} & 95.0 & \textbf{88.0} & \textbf{73.0} & 44.0 & 50.0 \\
        \rowcolor{gray!15} 
        \textbf{ANI (Ours)} & \textbf{100.0} & \textbf{98.0} & \textbf{88.0} & 67.0 & \textbf{68.0} & \textbf{82.0} \\
        \bottomrule
    \end{tabular}
    }
    \caption{\textbf{N-Digit Arithmetic Scaling.} Exact match accuracy (\%) across varying lengths of numerical sequences. The results demonstrate the catastrophic calculation failures of standard subword tokenization on longer sequences (e.g., 6+ digits) and highlight ANI's effectiveness as a structural stabilizer.}
    \label{tab:ndigit_scaling}
\end{table}

As shown in Table~\ref{tab:ndigit_scaling}, Qwen3-8B completely collapses starting from just 6 digits (6.0\%), proving that inherent LLM representations are highly fragile for longer inputs. Furthermore, even though standard SFT is trained on the same dataset, it falls significantly behind ANI as numbers extend to 10-12 digits (e.g., SFT's 50.0\% vs. ANI's 82.0\% at 12 digits) due to cumulative information loss from token fragmentation. This confirms that by injecting a mathematically unified, single-token topology, ANI acts as a structural stabilizer that successfully bypasses this bottleneck.

\paragraph{Optimal Injection Depth and Contextual Maturity.} 
To identify the most effective stage for numerical injection, we analyze the model's internal capability to disambiguate numerical roles across its layers. Figure~\ref{fig:layer_sensitivity} shows that the ability to distinguish between nominal identifiers and quantitative operands exhibits a distinct progression, reaching its absolute peak of 93\% at Layer 24. This peak indicates a state of maximum contextual maturity, where the hidden representations have fully integrated the surrounding linguistic context to resolve the functional ambiguity of numerical tokens. 

The impact of this contextual maturity is directly reflected in downstream reasoning performance. Our layer sensitivity analysis on the MATH benchmark shows a strong correlation with the linear probe results, where injecting the ANI module at Layer 24 achieves the highest accuracy of 58.76\%. Conversely, the early probing peak at Layer 3 fails to translate into MATH improvements, as it relies on surface-level lexical cues rather than deep contextual synthesis. This confirms our hypothesis that numerical injection is most effective at the depth where the model has attained sufficient contextual maturity to interpret the role of numbers.

\begin{figure}[t]
    \centering
    \includegraphics[width=\columnwidth]{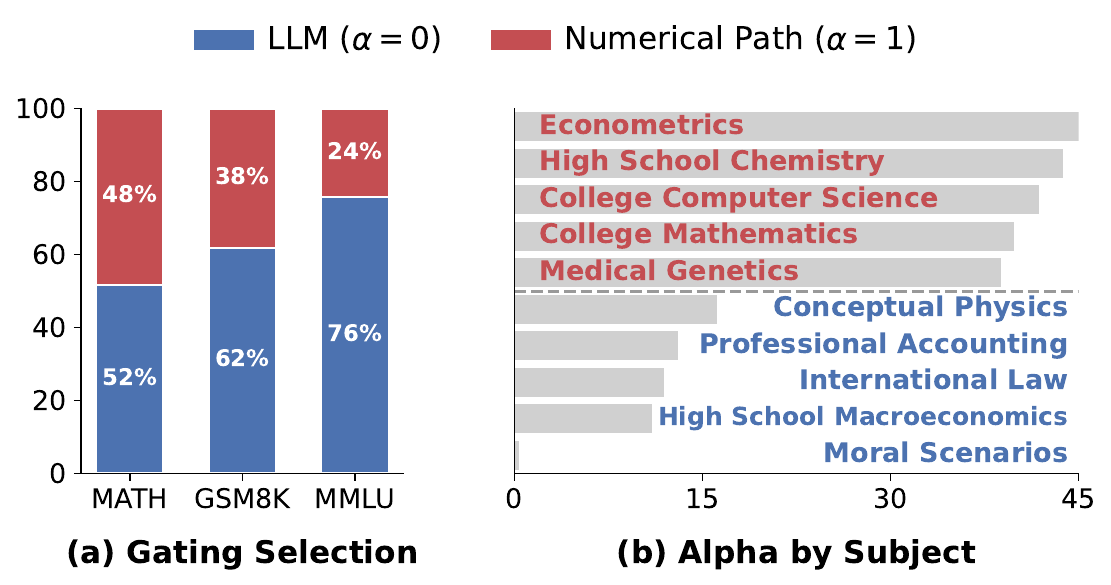}
    \vspace{-1.8em}
    \caption{\textbf{Interpretability Analysis of the Adaptive Gating Mechanism.} (a) Selection ratio across benchmarks (y-axis: proportion). (b) Selection ratio across MMLU subjects (x-axis: proportion of $\alpha=1$). }
    \label{fig:gating_analysis}
\end{figure}

\paragraph{Gating Behavior Analysis.}
To interpret the internal decision-making process of ANI, we analyze the gating network’s activation patterns across diverse reasoning tasks. As shown in Figure 6(a), the mechanism exhibits task-aware selectivity: for the calculation-heavy MATH benchmark, it activates numerical embeddings ($\alpha = 1$) in 48\% of instances, whereas this ratio drops to 24\% for the predominantly linguistic MMLU. The subject-level granularity in Figure 6(b) reveals a distinct semantic divide. Within the MMLU benchmark, the gating network selects the numerical path ($\alpha = 1$) significantly more often (over 35\% of cases) for STEM disciplines like Econometrics and Chemistry, while the selection ratio remains near-zero for Humanities such as International Law. Notably, even within STEM, theoretical subjects like Conceptual Physics show a lower activation frequency compared to calculation-intensive ones. This fine-grained discrimination confirms that ANI understands the functional necessity of numbers, ensuring that arithmetic reinforcement is triggered only when the semantic context demands it.

\paragraph{Impact of the Gating Mechanism.}
To dissect ANI's architectural components, we evaluate the necessity of context-aware gating and our sparse injection strategy (Table~\ref{tab:ablation}) against two baselines: (i) \textit{Static Injection} (no gating) and (ii) \textit{Digit-wise Injection} (per-token gating). Without gating, Static Injection suffers a severe collapse in reasoning (8.60\% MATH) and general capabilities (71.41\% MMLU), confirming that forcing numerical topology without context acts as disruptive noise. Furthermore, while Digit-wise Injection achieves comparable arithmetic performance, it remains suboptimal in preserving linguistic integrity (72.29\% MMLU). This indicates that broadcasting embeddings to every sub-token induces redundant signal interference that overwhelms the local semantic context. In contrast, ANI’s sparse intervention at the terminal token bottleneck successfully maximizes mathematical precision while effectively safeguarding the model's linguistic manifold.

\paragraph{Gating Accuracy and Tool-Routing Synergy.}
To evaluate the utility of our gating mechanism, we analyze its behavior across both real-world linguistic contexts and complex agentic workflows. First, on the FiNER-139 dataset~\cite{loukas2022finer}, which provides explicit nominal ground truth (e.g., "Debt Instrument Maturity Date"), our gating network achieves a 75.8\% precision in bypassing these nominal identifiers. This provides strong evidence that ANI effectively safeguards the model's linguistic manifold from semantic erasure. 
Furthermore, we evaluate its tool-use applicability on a highly deceptive synthetic dataset (detailed in Appendix~\ref{app:syn}) that mixes nominal traps with quantitative operands. While the baseline pre-trained model avoids passing nominal traps into tool arguments in only 69.6\% of cases, ANI achieves a 92.9\% success rate by precisely suppressing the nominal embeddings ($\alpha_{num}=0$). Together, these results confirm that ANI functions not merely as an arithmetic booster, but as a stable foundational architecture for reliable tool utilization.


\begin{table}[t]
    \centering
    \resizebox{\linewidth}{!}{
    \begin{tabular}{lccc}
        \toprule
        \textbf{Method} & \textbf{MATH} & \textbf{GSM8K} & \textbf{MMLU} \\
        \midrule
        Static Injection (No Gating) & 8.60 & 62.70 & 71.41 \\
        Digit-wise Injection & 58.70 & 90.37 & 72.29 \\
        \rowcolor{gray!15}
        \textbf{ANI (Ours)} & 58.76 & 90.14 & 72.67 \\
        \bottomrule
    \end{tabular}
    }
    \caption{\textbf{Ablation Study.} Impact of the gating mechanism and injection locus. Static Injection replaces tokens without gating, while Digit-wise Injection applies the gate to every sub-word digit token.}
    \label{tab:ablation}
\end{table}

\subsection{Generalization and Scalability} \label{sec:generalization}
To verify whether ANI represents a universally applicable enhancement, we evaluate its performance across diverse architectures and parameter scales, as summarized in Table~\ref{tab:scalability_generalization}. 

\paragraph{Architectural Robustness.} ANI demonstrates robust generalization across the Llama-3.1 and Mistral families. Crucially, the input-level baseline xVal routinely regresses below the standard SFT baseline across these diverse architectures. This regression clearly highlights the instability of context-agnostic substitution. In contrast, ANI effectively avoids this cascading degradation. While ANI achieves a 2.0\% gain on GSM8K for Mistral-7B, the slight regression in Mistral’s MATH suggests that late-stage injection at layer 30 (of 32) lacks sufficient residual depth for complex reasoning. Nevertheless, the overall trend confirms the framework’s cross-architecture portability. Optimal layers are detailed in Appendix~\ref{sec:hyper}.

\paragraph{Scaling across Model Sizes.}
The effectiveness of numerical topology injection is further validated across the Qwen family, ranging from 4B to 32B parameters. Notably, on the compact Qwen3-4B, ANI achieves accuracy improvements of 1.3 points on MATH and 2.4 points on GSM8K over the SFT baseline, demonstrating that even capacity-constrained models benefit from explicitly injected numerical representations. Furthermore, while standard fine-tuning and xVal suffer significant performance degradation on the 4B's GSM8K task compared to the pre-trained model, ANI effectively mitigates this loss, preserving robust reasoning capabilities. This consistent advantage extends to the larger Qwen3-32B model, where ANI continues to deliver the highest accuracy (63.1\% on MATH and 93.5\% on GSM8K), outperforming both the SFT baseline and xVal. This positive trajectory confirms that ANI provides reliable reasoning reinforcement as the backbone’s inherent capacity scales.

\begin{table}[t]
    \centering
    \resizebox{\linewidth}{!}{
    \begin{tabular}{llcc}
        \toprule
        \textbf{Model Family / Size} & \textbf{Method} & \textbf{MATH} & \textbf{GSM8K} \\
        \midrule
        \multirow{4}{*}{\textbf{Llama-3.1-8B}} 
        & Pre-trained & 20.5 & 55.3 \\
        & Baseline (SFT) & 26.4 & \textbf{63.5} \\
        & XVal & 21.2 & 59.5 \\
        \rowcolor{gray!15} \cellcolor{white}
        & \textbf{ANI (Ours)} & \textbf{27.2} & \textbf{63.5} \\
        \midrule
        \multirow{4}{*}{\textbf{Mistral-7B-v0.3}} 
        & Pre-trained & 11.4 & 37.5 \\
        & Baseline (SFT) & \textbf{16.6} & 52.4 \\
        & XVal & 16.2 &  51.0 \\
        \rowcolor{gray!15} \cellcolor{white}
        & \textbf{ANI (Ours)} & 15.0 & \textbf{54.4} \\
        \specialrule{.08em}{.5ex}{.5ex} 
        \multirow{4}{*}{\textbf{Qwen3-4B}} 
        & Pre-trained & 47.0 & \textbf{87.4} \\ 
        & Baseline (SFT) & 48.2 & 82.8 \\
        & XVal & 47.2 & 82.5 \\
        \rowcolor{gray!15} \cellcolor{white}
        & \textbf{ANI (Ours)} & \textbf{49.5} & 85.2 \\
        \midrule
        \multirow{4}{*}{\textbf{Qwen3-32B}} 
        & Pre-trained & 60.0 & 91.4 \\
        & Baseline (SFT) & 62.5 & 92.0 \\
        & XVal & 61.6 & 89.2 \\
        \rowcolor{gray!15} \cellcolor{white}
        & \textbf{ANI (Ours)} & \textbf{63.1} & \textbf{93.5} \\
        \bottomrule
    \end{tabular}
    }
    \caption{\textbf{Scalability and Generalization Results.} Comparison of mathematical reasoning performance across various distinct architectures and model sizes.}
    \label{tab:scalability_generalization}
\end{table}

\section{Conclusion}
In this work, we addressed the fundamental tension between discrete subword tokenization and precise numerical reasoning in Large Language Models. We introduced ANI (Adaptive Numerical Injection), a hybrid framework that governs the selective activation of specialized numerical embeddings based on the linguistic context. By employing a context-aware gating mechanism, ANI effectively resolves the functional ambiguity of numbers, selectively enriching quantitative operands with mathematical topology while safeguarding nominal identifiers to prevent semantic erasure.

Our empirical results on the Qwen3-8B and various model families demonstrate that ANI significantly bolsters arithmetic reasoning on complex benchmarks while maintaining the model’s pre-trained linguistic manifold. These findings confirm that the adaptive integration of numerical information is a key factor in unifying arithmetic precision with semantic fluency in LLMs.

\section{Limitations}
Despite strong empirical gains, several aspects of our framework remain open for further refinement.

First, while ANI is designed to be model-agnostic, its current implementation inherits the representational resolution of the underlying numerical encoder. Although our present configuration of 10 integer and 6 fractional digits is sufficient for most benchmarks, it may not generalize to specialized domains, such as science, requiring extreme high precision.

Second, our method injects numerical features at a layer identified via linear probing as the point of peak \textit{contextual maturity}. While we observe this peak to be stable within model families, developing an automated, architecture-agnostic protocol for identifying this layer would further improve portability across emerging LLM architectures. 

Finally, ANI introduces a lightweight gating and projection module during the forward pass. While negligible relative to the backbone model's compute footprint, further optimization may be necessary for extreme cases of highly latency-sensitive deployments. 

\section*{Acknowledgments}
This work was supported by Institute of Information \& communications Technology Planning \& Evaluation (IITP) grant funded by the Korea government (MSIT) (No. 2022-0-00077/RS-2022-II220077, AI Technology Development for Commonsense Extraction, Reasoning, and Inference from Heterogeneous Data).

\bibliography{custom}
\clearpage 

\appendix

\section{Mathematical Derivation of the Binarization Penalty}
\label{sec:Differentiable}
The core innovation of ANI lies in its ability to perform discrete functional selection between nominal identifiers and quantitative operands while remaining end-to-end trainable. Since traditional discrete gating based on a Bernoulli distribution is non-differentiable, we utilize a continuous sigmoid relaxation coupled with an explicit binarization penalty.

\subsection{Continuous Relaxation and Binarization}
Our gating MLP computes a continuous probability based on the target logit $z_{num}$:
\begin{equation}
\alpha_{num} = \sigma \big( MLP_{gate}([\mathbf{h}^{(l)}_{t_{L}}; \mathbf{e}_{num}]) \big) \in (0, 1)
\end{equation}
To induce discreteness and force the continuous scale $\alpha_{num}$ toward a deterministic $0$ or $1$, we introduce a regularization term to the objective function:
\begin{equation}
\mathcal{L}_{gate} = \lambda \mathbb{E}[\alpha_{num}(1-\alpha_{num})]
\end{equation}
This parabolic penalty reaches its maximum at $\alpha_{num} = 0.5$ and minimizes at the bounds $0$ and $1$. By incorporating this penalty into the standard Next-Token Prediction loss, the network is effectively pushed away from ambiguous intermediate states during training, acting as a soft surrogate for discrete step functions.

\subsection{Deterministic Inference}
During inference, we require absolute zero-interference for nominal identifiers to prevent semantic erasure. We achieve this by applying a hard threshold directly to the continuous activation without stochastic noise:
\begin{equation}
\alpha_{num} = \mathbb{I}(\sigma(z_{num}) > 0.5)
\end{equation}
This formulation allows ANI to maintain a strict, deterministic 'on/off' switching regime during inference while leveraging stable, continuous gradients to optimize the Gating Network during training, bypassing the instability commonly associated with stochastic estimators.

\begin{algorithm}[t]
    \caption{Forward Pass of ANI}
    \label{alg:ani_forward}
    
    {\bfseries Input:} Token sequence $S = \{t_1, \dots, t_N\}$, 
                       Set of numerical entities $V = \{(v_k, \text{idx}_k)\}_{k=1}^M$,
                       Target Layer $l$, Mode \texttt{is\_training} \\
    {\bfseries Parameters:} $\mathbf{W}_P$ (Linear Projection), $\theta_{\text{gate}}$ (Gating MLP), FoNE periods $P$
    
    \begin{algorithmic}[1]
       \STATE $\mathbf{H}^{(l)} \leftarrow \text{LLM}_{1:l}(S)$ \quad \hfill \textit{// 1. Contextual Representation Extraction}
       
       \FOR{each $(v_k, \text{idx}_k) \in V$}
          \STATE $i \leftarrow \text{idx}_k$ \quad \hfill \textit{// Identify terminal sub-token index}
          
          \STATE $\phi(v_k) \leftarrow \text{FoNE}(v_k; P)$ \quad \hfill \textit{// 2. Topology Encoding}
          \STATE $\mathbf{e}_{num} \leftarrow \mathbf{W}_P \cdot \phi(v_k)$ \quad \hfill \textit{// Manifold Alignment}
          
          \STATE $z_{num} \leftarrow MLP_{\text{gate}}([\mathbf{h}_i^{(l)}; \mathbf{e}_{num}]; \theta_{\text{gate}})$ \quad \hfill \textit{// 3. Semantic Probing}
          \IF{\texttt{is\_training}}
              \STATE $\alpha_{num} \leftarrow \sigma(z_{num})$ \quad \hfill \textit{// Continuous scale for binarization penalty}
          \ELSE
              \STATE $\alpha_{num} \leftarrow \mathbb{I}(\sigma(z_{num}) > 0.5)$ \quad \hfill \textit{// Deterministic hard selection}
          \ENDIF
          
          \STATE $\mathbf{h}_i^{(l)} \leftarrow \mathbf{h}_i^{(l)} + \alpha_{num} \cdot \mathbf{e}_{num}$ \quad \hfill \textit{// 4. Adaptive Integration}
       \ENDFOR
       
       \STATE $\text{Output} \leftarrow \text{LLM}_{l+1:L}(\mathbf{H}^{(l)})$ \quad \hfill \textit{// Residual Layers Processing}
       \STATE {\bfseries Return} Output
    \end{algorithmic}
\end{algorithm}

\section{Procedural Overview of ANI}
Algorithm~\ref{alg:ani_forward} details the plug-and-play intervention of the ANI module within the intermediate layers of a backbone LLM. To minimize signal redundancy and potential feature entanglement, the injection is strictly targeted at the terminal sub-token index ($idx_k$) of each numerical entity, which serves as an information bottleneck in decoder-only architectures.
\begin{enumerate}
\item \textbf{Topology Encoding (Lines 4-5):} Raw scalars are mapped to a high-dimensional mathematical manifold via FoNE and subsequently aligned to the model's latent space using a learnable Linear Projection ($W_P$).
\item \textbf{Context-Aware Gating (Lines 6-11):} To determine arithmetic necessity, the MLP outputs a raw logit that is mapped via a standard sigmoid function. During training, the gate acts as a continuous scaler ($\alpha_{num} \in (0, 1)$), allowing end-to-end optimization guided by an explicit binarization penalty. During inference, it applies a strict deterministic threshold ($\alpha_{num} \in \{0, 1\}$) to ensure absolute zero-interference for nominal identifiers.
\item \textbf{Efficiency:} Tokens not identified as numerical bypass the module entirely, ensuring zero interference with the model's pre-trained linguistic knowledge.
\end{enumerate}

\section{Implementation Details of ANI}
\label{sec:detailed}

\subsection{Experimental setup}
\label{sec:experimentalenvironments}
We run our experiments on a machine equipped with Intel Xeon(R) Gold 6226R CPUs and Nvidia RTX 3090/A6000 GPUs.
We implement ANI using Python 3.10, PyTorch 2.1, Hugging Face Transformers, and vLLM.

\subsection{Datasets and Evaluation Protocols}
To evaluate the numerical reasoning and general linguistic stability of ANI, we employ specific few-shot Chain-of-Thought (CoT) protocols across three primary benchmarks:
\begin{itemize}
    \item \textbf{MATH (4-shot CoT):} To assess complex multi-step mathematical reasoning, we use a 4-shot prompt where the model is instructed to provide step-by-step solutions and encapsulate the final answer within a \texttt{\textbackslash boxed\{\}} command.
    \item \textbf{GSM8K (8-shot CoT):} For grade-school word problems, an 8-shot CoT protocol is utilized. The model is required to provide the final numerical value following the \texttt{\#\#\#\#} delimiter to ensure consistent answer extraction.
    \item \textbf{MMLU (5-shot):} To monitor general-purpose linguistic knowledge and detect potential semantic erasure, we evaluate the model across 57 subjects using a standard 5-shot setting. Evaluation is performed by comparing the log-probabilities of the candidate options (A, B, C, D).
\end{itemize}

\subsection{Diagnostic Dataset Construction}
The \textbf{Gating Network} is trained on a composite diagnostic dataset designed to help the model distinguish between the functional roles of numbers (Nominal vs. Quantitative).

\begin{itemize}
    \item \textbf{Quantitative Samples:} 10,000 samples are extracted from \textbf{OpenMathInstruct-2}, focusing on contexts where numerical values serve as operands for arithmetic operations. Crucially, the OpenMathInstruct-2 dataset has already undergone a rigorous LLM-based decontamination pipeline to remove any potential paraphrases of the MATH and GSM8K evaluation sets~\cite{toshniwal2024openmathinstruct}. This inherently precludes any risk of data leakage into our diagnostic training phase.
    \item \textbf{Nominal Samples:} 10,000 samples are curated from the \textbf{Magpie (Pro-MT-300K)} dataset, representing linguistic contexts where numbers act as identifiers.
    \item \textbf{Filtering Strategy:} To ensure the purity of nominal samples, we apply a strict keyword-based filter. Any Magpie entry containing math-related terms (e.g., \textit{math, calculation, algebra}) or programming-related terms (e.g., \textit{python, algorithm, sql}) is excluded to prevent the leakage of quantitative reasoning signals into the nominal set.
\end{itemize}

\subsection{Baselines}
\begin{itemize}
    \item \textbf{Reference:} The official pre-trained \texttt{Qwen3-8B} model without any additional fine-tuning, serving as the lower-bound performance metric.
    \item \textbf{SFT Baseline:} A standard supervised fine-tuning approach that utilizes the same composite dataset but relies exclusively on vanilla subword tokenization without specialized numerical representations.
    \item \textbf{Digit-wise (Additive):} Implements an additive fusion scheme where numerical features are summed with textual embeddings at the input layer ($h_{num} \leftarrow h_{num} + e_{num}$). This baseline evaluates the impact of simple feature augmentation without the benefit of context-aware gating.
    \item \textbf{xVal~\cite{golkar2023xval} (Substitution)}: This method addresses numerical fragmentation by mapping scalar values to a continuous, single-token representation via multiplicative magnitude encoding in a learnable direction. It is designed to capture quantitative magnitude and continuity directly within the embedding space. However, because it relies on context-agnostic substitution (replacing all numbers with a generic \texttt{[NUM]} token) at the input layer, it often fails to preserve the semantic role of numbers when they function as nominal identifiers.
    \item \textbf{FoNE~\cite{zhou2025fone} (Substitution)}: This approach utilizes Fourier features to provide precise single-token number embeddings by representing residues in various modular groups. By mapping raw scalars through periodic sine and cosine activation functions, it effectively captures both magnitude and fine-grained fractional details in a high-dimensional feature vector. Similar to xVal, its global substitution strategy ($h_{num} \leftarrow e_{num}$) ensures arithmetic precision but triggers semantic erasure, stripping numbers of their linguistic identity in multi-modal contexts.
    \item \textbf{NumeroLogic~\cite{schwartz2024numerologic} (In-context)}: A representation reformatting approach that enhances reasoning by prepending the digit count to each number (e.g., \texttt{"2:42"}), providing explicit place-value information directly in the input sequence. This structural encoding acts as a simplified Chain-of-Thought (CoT), prompting the model to reason about magnitude before generation. Unlike continuous methods, it operates entirely within the discrete textual space via text pre-processing, focusing on improving the model's inherent interpretation of decimal strings.
\end{itemize}

\subsection{Evaluation Prompts}
\label{app:evalprompt}
\paragraph{MATH Evaluation Template}

\begin{quote}
\ttfamily
{[System]} \\
You are a helpful math reasoning assistant. Please reason step by step. \\
\\
{[User]} \\
Solve the following math problem step-by-step. Put your final answer in \textbackslash boxed\{\}. \\
\\
Below are some examples: \\
Question: \\
Olivia has \$23. She bought five bagels for \$3 each. How much money does she have left? \\
My solution: \\
First, calculate the total cost of the bagels. Olivia bought 5 bagels at \$3 each. Cost = 5 * 3 = 15. She started with \$23 and spent \$15. Money left = 23 - 15 = 8. The final answer is: \textbackslash boxed\{8\} \\
Question: \\
A pen and its ink refill together cost \$1.10. The pen costs \$1 more than the ink refill. What is the cost of the pen in dollars? \\
My solution: \\
Let p be the cost of the pen and i be the cost of the ink refill. We are given: 1. p + i = 1.10, 2. p = i + 1. Substitute equation 2 into equation 1: (i + 1) + i = 1.10 -> 2i + 1 = 1.10 -> 2i = 0.10 -> i = 0.05. Now find p: p = i + 1 = 0.05 + 1 = 1.05. The final answer is: \textbackslash boxed\{1.05\} \\
Question: \\
What is the value of f(f(1)) if f(x) = 2x + 3? \\
My solution: \\
First, we calculate f(1): f(1) = 2(1) + 3 = 2 + 3 = 5. Now substitute this into f(x): f(f(1)) = f(5) = 2(5) + 3 = 10 + 3 = 13. The final answer is: \textbackslash boxed\{13\} \\
Question: \\
How many ways can 3 people (Alice, Bob, Charlie) line up in a single file line? \\
My solution: \\
This is a permutation problem. For the first spot, there are 3 choices. For the second, 2 remaining. For the last, 1 choice left. Total ways = 3 * 2 * 1 = 6. The final answer is: \textbackslash boxed\{6\} \\
Now, solve this problem: \\
Question: \{Target Problem\}
\end{quote}

\paragraph{GSM8K Evaluation Template}

\begin{quote}
\ttfamily
{[System]} \\
You are a helpful math assistant. Please reason step by step. \\
\\
{[User]} \\
Solve the following math problem step-by-step. At the end of your response, provide the final answer in the format '\#\#\#\# [value]'. \\
\\
Below are some examples of math problems: \\
Q: There are 15 trees in the grove. Grove workers will plant trees in the grove today. After they are done, there will be 21 trees. How many trees did the grove workers plant today? \\
A: There are 15 trees originally. Then there were 21 trees after the Grove workers planted some more. So there must have been 21 - 15 = 6. The answer is 6. \\
Q: If there are 3 cars in the parking lot and 2 more cars arrive, how many cars are in the parking lot? \\
A: There are 3 cars originally. 2 more cars arrive. 3 + 2 = 5. The answer is 5. \\
Q: Leah had 32 chocolates and her sister had 42. If they ate 35, how many pieces do they have left in total? \\
A: Leah had 32 chocolates and her sister had 42. That means there were 32 + 42 = 74 chocolates. 35 were eaten. So in total they still have 74 - 35 = 39 chocolates. The answer is 39. \\
Q: Jason had 20 lollipops. He gave Denny some lollipops. Now Jason has 12 lollipops. How many lollipops did Jason give to Denny? \\
A: Jason had 20 lollipops. Since he only has 12 now, he must have given the rest to Denny. The number of lollipops he gave to Denny must have been 20 - 12 = 8. The answer is 8. \\
Q: Shawn has five toys. For Christmas, he got two toys each from his mom and dad. How many toys does he have now? \\
A: He has 5 toys. As he got 2 from mom and 2 from dad, he got 2 + 2 = 4 more. 5 + 4 = 9. The answer is 9. \\
Q: There were 9 computers in the server room. Five more computers were installed each day for 4 days. How many computers are now in the server room? \\
A: There were 9 computers. Then 5 more were installed for 4 days, which means 5 * 4 = 20 computers were added. 9 + 20 = 29. The answer is 29. \\
Q: Michael had 58 golf balls. On tuesday, he lost 23 golf balls. On wednesday, he lost 2 more. How many golf balls did he have at the end of wednesday? \\
A: Michael initially had 58 balls. He lost 23 on Tuesday, so after that he had 58 - 23 = 35 balls. On Wednesday he lost 2 more so now he has 35 - 2 = 33 balls. The answer is 33. \\
Q: Olivia has \$23. She bought five bagels for \$3 each. How much money does she have left? \\
A: She bought 5 bagels for \$3 each. This means she spent 5 * 3 = \$15. She had \$23 so she has 23 - 15 = 8. The answer is 8. \\
Now, solve this problem: \\
Q: \{Target Question\} \\
A:
\end{quote}

\paragraph{MMLU Evaluation Template}
\begin{quote}
\textit{[System]} \\
The following are multiple choice questions (with answers) about \texttt{\{subject\}}.
\textit{[User]} \\
Question: \{Few-shot Example 1\} \\
(A) \{Option 1\} (B) \{Option 2\} (C) \{Option 3\} (D) \{Option 4\} \\
Answer: A

...

Question: \{Few-shot Example 5\} \\
(A) \{Option 1\} (B) \{Option 2\} (C) \{Option 3\} (D) \{Option 4\} \\
Answer: C

Question: \{Target Question\} \\
(A) \{Option 1\} (B) \{Option 2\} (C) \{Option 3\} (D) \{Option 4\} \\
Answer:
\end{quote}

\subsection{Hyperparameters}
\label{sec:hyper}
To ensure a fair and rigorous evaluation of ANI, we maintain a consistent set of hyperparameters across diverse model architectures and scales. All models are evaluated with a maximum sequence length of 4,096 tokens. To ensure high numerical resolution, the integer and fractional lengths for numerical representations are fixed at 10 and 6 digits, respectively.

The Gating Network consists of a 2-layer MLP with a dropout rate of 0.1 to prevent overfitting during the specialized training phase. The optimal injection layer index ($l$) is determined based on the contextual maturity of each model. For Qwen3-8B, Layer 24 is selected based on our diagnostic linear probing analysis. For other architectures, including Qwen3-4B, Qwen3-32B, Llama-3.1-8B, and Mistral-7B, the injection layers are set to 21, 37, 14, and 30, respectively, determined through similar probing heuristics or empirical validation.

All models are optimized using the QLoRA framework with a rank ($r$) of 64 and an alpha ($\alpha$) of 128. We employ the AdamW optimizer with a constant learning rate of $5 \times 10^{-5}$, utilizing a per-device batch size of 1 and a gradient accumulation of 32 steps to ensure stable convergence. 

\begin{figure}[t]
    \centering
    \includegraphics[width=0.87\linewidth]{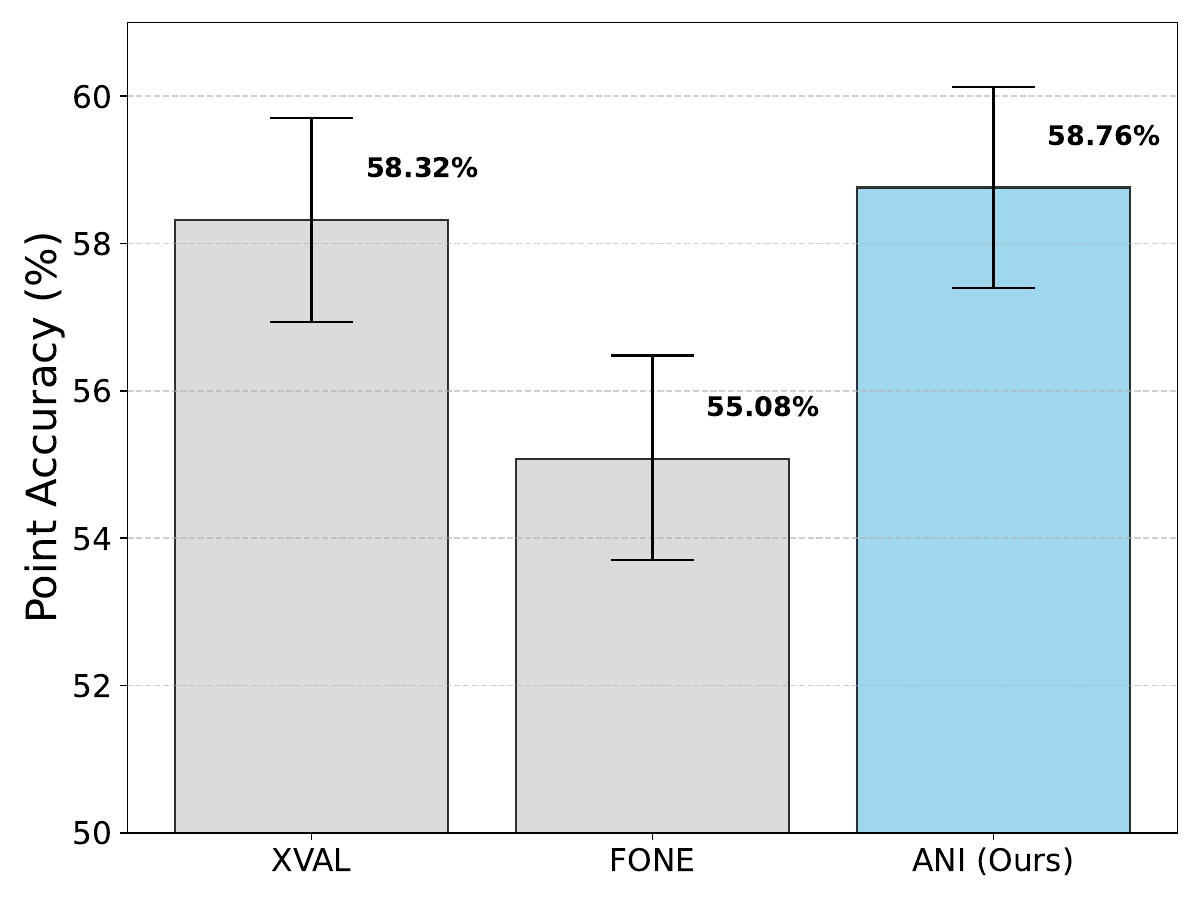}
    \caption{\textbf{Model performance comparison with 95\% CI.} Bars represent the overall accuracy on the MATH benchmark, with error bars denoting the 95\% confidence intervals derived from paired bootstrap resampling.}
    \label{fig:perf_comparison_ci} 
    \vspace{-1.5em}
\end{figure}

\section{Additional Experiment Results}
\subsection{Confidence Intervals and Difficulty-wise Analysis}
\label{app:confidence_intervals}
To evaluate the statistical reliability of our empirical results and ensure that the performance gains are not artifacts of random variance, we employ paired bootstrap resampling on the evaluation outputs. Specifically, we draw $1,000$ bootstrap samples from the test sets of the MATH benchmark ($N = 5,000$ total; Levels 1-3: $N = 2,462$, Levels 4-5: $N = 2,538$) and the MMLU benchmark ($N = 14,042$). We then compute the accuracy delta ($\Delta$) between ANI and the respective baselines for each sample to determine the $2.5^{th}$ and $97.5^{th}$ percentiles as the empirical bounds. 

First, we report the statistical reliability compared against all input-level baselines. As illustrated in Figure~\ref{fig:perf_comparison_ci}, the 95\% confidence interval (CI) analysis across all input-level variations reveals that ANI consistently achieves the highest lower and upper empirical bounds compared to all other baselines, thereby demonstrating the stability of our approach.

Furthermore, when diving deeper into a granular, difficulty-wise analysis against xVal (Table~\ref{tab:math_difficulty}), this statistical advantage becomes even more pronounced. On simpler segments (Levels 1, 2, and 3), the performance difference is marginal with an accuracy delta of $-0.0049$ and a paired bootstrap 95\% CI that includes zero ($[-0.0150, +0.0057]$), resulting in a non-significant $p$-value ($p = 0.8303$). Conversely, in the most challenging intervals (Levels 4 and 5), ANI exhibits a distinct performance improvement, with an accuracy delta of $+0.0134$. Crucially, this high-difficulty confidence interval strictly excludes zero ($[+0.0004, +0.0268]$), yielding a statistically significant $p$-value of less than $0.05$ ($p = 0.0241$). This fine-grained evaluation confirms that ANI’s performance is both meaningful and effective in complex arithmetic reasoning compared to input-level baselines.

Finally, our bootstrap analysis on MMLU confirms that ANI (72.67\%) significantly mitigates semantic drift compared to the standard fine-tuning baseline (72.24\%). The accuracy delta of $+0.0043$ yields a paired bootstrap 95\% CI that strictly excludes zero ($[+0.0017, +0.0069]$, $p=0.0008$), proving its robustness in preserving general linguistic capabilities alongside mathematical improvements.


\begin{table}[t]
\begin{center}
\begin{small}
\begin{sc}
\resizebox{\columnwidth}{!}{%
\begin{tabular}{lccc}
\hline
Group & Acc. Diff & Paired BS 95\% CI & BS $p$ (one-sided) \\
\hline
Level 1, 2, 3 & $-0.0049$ & $[-0.0150, +0.0057]$ & $0.8303$ \\
Level 4, 5     & $+0.0134$ & $[+0.0004, +0.0268]$ & $0.0241^{*}$ \\
\hline
\multicolumn{4}{r}{\scriptsize{$^*p < 0.05$}}
\end{tabular}%
}
\end{sc}
\end{small}
\end{center}
\vspace{-1em}
\caption{\textbf{Fine-grained bootstrap statistical analysis.} Comparison of ANI's accuracy gain relative to xVal on the MATH benchmark.}
\label{tab:math_difficulty}
\vspace{-1em}
\end{table}




\subsection{Synthetic Evaluation on Gating Accuracy and Agentic Tool-Routing}
\label{app:syn}
To rigorously evaluate our internal gating mechanism against ground-truth answers where specific numerical values are explicitly irrelevant, we constructed a synthetic evaluation set comprising 50 highly deceptive queries. These queries intentionally deploy nominal identifiers (e.g., ``Employee 2048'') as traps alongside actual quantitative operands (e.g., ``earns \$5,000'' and ``10\% bonus''). This setup tests whether the model's gating can correctly align with the ground truth and bypass nominal traps, precisely routing only the actual arithmetic operands to external tools (e.g., executing \texttt{calculator(["5000", "0.10"])}).

The experimental results demonstrate the robustness of ANI's self-filtering capability. While the baseline pre-trained model avoids passing nominal traps into tool arguments in only 69.6\% of cases, ANI achieves a 92.9\% success rate. This substantial reduction in argument routing errors provides concrete evidence that ANI functions as a robust foundational architecture, ensuring representation integrity and reliable tool utilization within complex agent workflows.




\section{Use of AI Assistants}
In this paper, we utilized Gemini, a large-scale language model developed by Google, to refine English expressions and enhance linguistic clarity.

\end{document}